\documentclass[5pt]{article}

\usepackage[letterpaper]{geometry}
\usepackage{spconf,amsmath,epsfig}
\usepackage{enumitem}

\usepackage{cite}
\usepackage{framed,multirow,booktabs}
\usepackage{graphicx}
\usepackage{amssymb,bm,amsfonts}
\usepackage{algorithmic}
\usepackage{textcomp}
\usepackage{amsthm}
\usepackage{lineno}
\usepackage{caption}

\let\temp\footnote
\renewcommand \footnote[1]{\temp{\ninept#1}}

\usepackage{fancyhdr}
\begin{document}\sloppy
\topmargin=0mm

\def\x{{\mathbf x}}
\def\L{{\cal L}}

\title{Radical analysis network for zero-shot learning in printed Chinese character recognition}
%
\name{Jianshu Zhang, Yixing Zhu, Jun Du and Lirong Dai}
\address{National Engineering Laboratory for Speech and Language Information Processing\\
University of Science and Technology of China,
Hefei, Anhui, P. R. China\\ xysszjs@mail.ustc.edu.cn, zyxsa@mail.ustc.edu.cn, jundu@ustc.edu.cn, lrdai@ustc.edu.cn}

\maketitle

\begin{abstract}
  Chinese characters have a huge set of character categories, more than 20,000 and the number is still increasing as more and more novel characters continue being created. However, the enormous characters can be decomposed into a compact set of about 500 fundamental and structural radicals. This paper introduces a novel radical analysis network (RAN) to recognize printed Chinese characters by identifying radicals and analyzing two-dimensional spatial structures among them. The proposed RAN first extracts visual features from input by employing convolutional neural networks as an encoder. Then a decoder based on recurrent neural networks is employed, aiming at generating captions of Chinese characters by detecting radicals and two-dimensional structures through a spatial attention mechanism. The manner of treating a Chinese character as a composition of radicals rather than a single character class largely reduces the size of vocabulary and enables RAN to possess the ability of recognizing unseen Chinese character classes, namely zero-shot learning.
\end{abstract}
\begin{keywords}
Chinese characters, radical analysis, zero-shot learning, encoder-decoder, attention
\end{keywords}
\section{Introduction}
\label{sec:Introduction}
The recognition of Chinese characters is an intricate problem due to a large number of existing character categories (more than 20,000), increasing novel characters (e.g. the character ``Duang'' created by Jackie Chan) and complicated internal structures. However, most conventional approaches~\cite{liu2004online} can only recognize about 4,000 commonly used characters with no capability of handling unseen or newly created characters. Also each character sample is treated as a whole without considering the similarity and internal structures among different characters.

\begin{figure}
\centering
\includegraphics[width=3.4in]{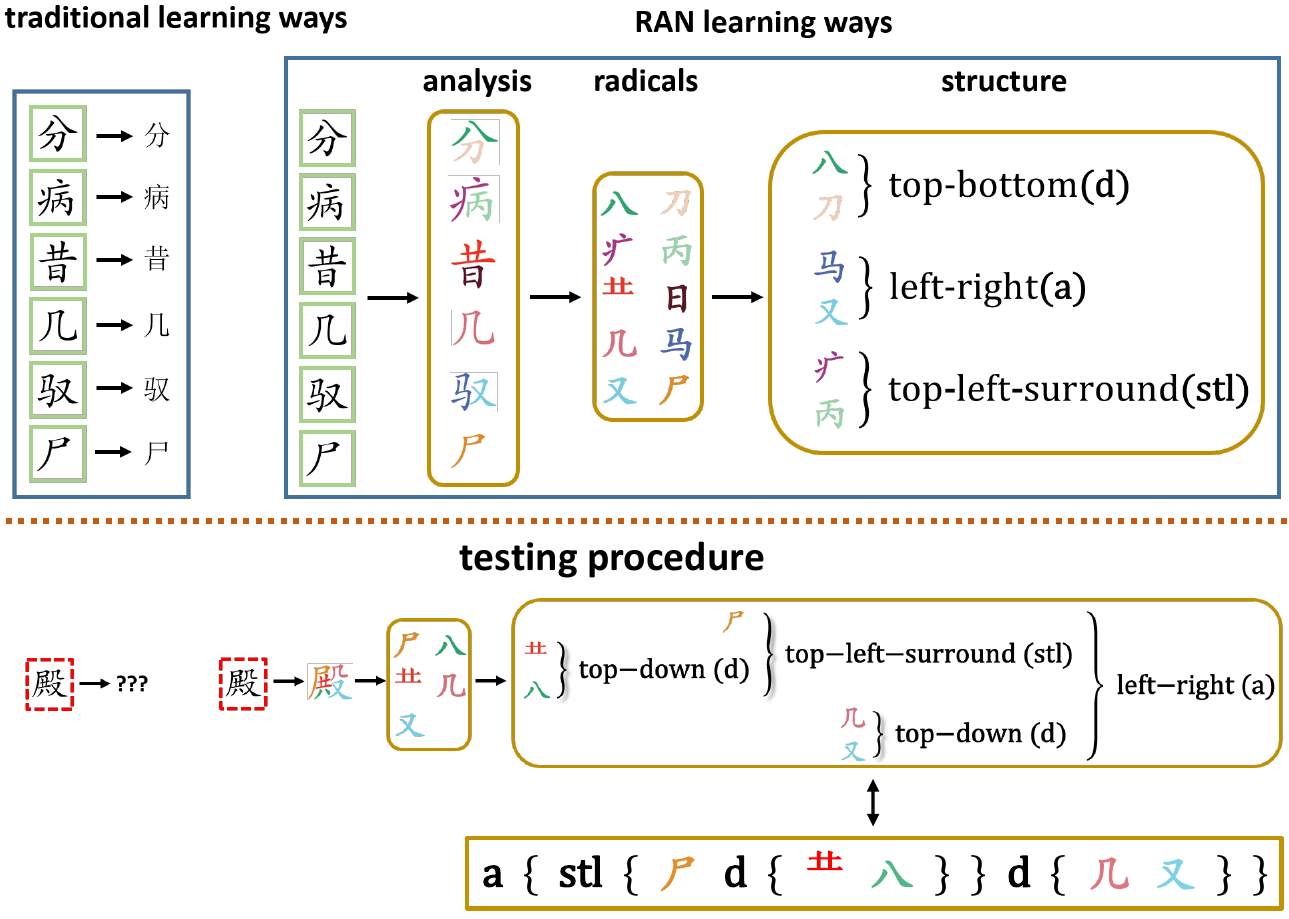}
\captionsetup{font=small}
\caption{The comparison between the conventional whole-character based methods and the proposed RAN method.}
\label{fig:RANdescription}
\end{figure}

It is well-known that all Chinese characters are composed of basic structural components, called radicals. Only about 500 radicals~\cite{li2013writing} are adequate to describe more than 20,000 Chinese characters. Therefore, it is an intuitive way to decompose Chinese characters into radicals and describe their spatial structures as captions for recognition. In this paper, we propose a novel radical analysis network (RAN) to generate captions for recognition of Chinese characters. RAN has two distinctive properties compared with traditional methods: 1) The size of the radical vocabulary is largely reduced compared with the character vocabulary; 2) It is a novel zero-shot learning of Chinese characters which can recognize unseen Chinese characters in the recognition stage because the corresponding radicals and spatial relationships are learned from other seen characters in the training stage. Fig.~\ref{fig:RANdescription} illustrates a clear comparison between traditional methods and RAN for recognizing Chinese characters. In the traditional methods, the classifier takes the character input as a single picture and tries to learn a mapping between the input picture and a pre-defined class. If the testing character class is unseen in training samples, like the character in red dashed rectangle in Fig.~\ref{fig:RANdescription}, it will be misclassified. While RAN attempts to imitate the manner of recognizing Chinese characters as Chinese learners. For example, before asking children to remember and recognize Chinese characters, Chinese teachers first teach them to identify radicals, understand the meaning of radicals and grasp the possible spatial structures between them. Such learning way is more generative and helps improve the memory ability of students to learn so many Chinese characters, which is perfectly adopted in RAN. As the example in top-right part of Fig.~\ref{fig:RANdescription} shows, the six training samples contain ten different radicals and exhibit top-bottom, left-right and top-left-surround spatial structures. When encountering the unseen character during testing, RAN can still generate the corresponding caption of that character (i.e. the bottom-right rectangle of Fig.~\ref{fig:RANdescription}) as it has already learned the essential radicals and structures. Technically, the enormous Chinese characters, as well as the newly created characters can all be identified by a compact set of radicals and spatial structures learned in the training stage.

In the past few decades, lots of efforts have been made for radical-based Chinese character recognition. \cite{ma2008new} over-segmented characters into candidate radicals and could only handle the left-right structure. \cite{wang2001optical} first detected separate radicals and then employed a hierarchical radical matching method to recognize a Chinese character. Recently, \cite{wang2017label} also tried to detect position-dependent radicals using a deep residual network with multi-labeled learning. Generally, these approaches have difficulties when dealing with the complicated 2D structures between radicals and do not focus on the recognition of unseen character classes.

Regarding to the network architecture, the proposed RAN is an improved version of the attention-based encoder-decoder model in~\cite{bahdanau2014neural}. We employ convolutional neural networks (CNN)~\cite{krizhevsky2012imagenet} as the encoder to extract high-level visual features from input Chinese characters. The decoder is recurrent neural networks with gated recurrent units (GRU)~\cite{chung2014empirical} which converts the high-level visual features into output character captions. We adopt a coverage based spatial attention model built in the decoder to detect the radicals and internal two-dimensional structures simultaneously. The GRU based decoder also performs like a potential language model aiming to grasp the rules of composing Chinese character captions after successfully detecting radicals and structures. The main contributions of this study are summarized as follows:
\begin{itemize}\setlength{\itemsep}{1pt}
  \item We propose RAN for the zero-shot learning of Chinese character recognition, solving the problem of handling unseen or newly created characters.
  \item We describe how to caption Chinese characters based on detailed analysis of Chinese radicals and structures.
  \item We experimentally demonstrate how RAN performs on recognizing seen/unseen Chinese characters and show its efficiency through attention visualization.
\end{itemize}


\section{Radical analysis}
\label{sec:Radical analysis}
Compared with enormous Chinese character categories, the amount of radical categories is quite limited. It is declared in GB13000.1 standard~\cite{li2013writing}, which is published by National Language Committee of China, that 20902 Chinese characters are composed of 560 different radicals.
\begin{figure}
\centering
\includegraphics[width=2.5in]{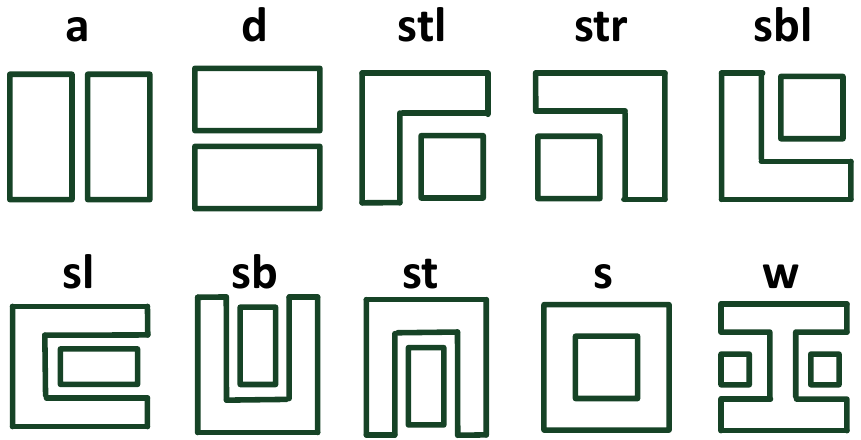}
\captionsetup{font=small}
\caption{Graphical representation of ten common spatial structures between Chinese radicals.}
\label{fig:StructureDescription}
\end{figure}
Following the strategy in cjk-decomp~\footnote{https://github.com/amake/cjk-decomp}, we decompose Chinese characters into corresponding captions. Regarding to spatial structures among radicals, we show ten common structures in Fig.~\ref{fig:StructureDescription} where ``\textbf{a}'' represents a left-right structure, ``\textbf{d}'' represents a top-bottom structure, ``\textbf{stl}'' represents a top-left-surround structure, ``\textbf{str}'' represents a top-right-surround structure, ``\textbf{sbl}'' represents a bottom-left-surround structure, ``\textbf{sl}'' represents a left-surround structure, ``\textbf{sb}'' represents a bottom-surround structure, ``\textbf{st}'' represents a top-surround structure, ``\textbf{s}'' represents a surround structure and ``\textbf{w}'' represents a within structure. 

We use a pair of braces to constrain a single structure in character caption. Taking ``stl'' as an example, it is captioned as ``stl \{ radical-1 radical-2 \}''. Usually, like the common instances mentioned above, a structure is described by two different radicals. However, as for some unique structures, they are described by three or more radicals. 

\section{Network architecture of RAN}
\label{sec:Network architecture of RAN}

The attention based encoder-decoder model first learns to encode input into high-level representations. A fixed-length context vector is then generated via weighted summing the high-level representations. The attention performs as the weighting coefficients so that it can choose the most relevant parts from the whole input for calculating the context vector. Finally, the decoder uses this context vector to generate variable-length output sequence word by word. The framework has been extensively applied to many applications including machine translation~\cite{cho2014learning}, image captioning~\cite{vinyals2015show,xu2015show}, video processing \cite{gao2017video} and handwriting recognition~\cite{zhang2017watch,zhang2017gru,zhang2018multi}.

\subsection{CNN encoder}
\label{sec:CNN encoder}
\begin{figure}
\centering
\includegraphics[width=3.3in]{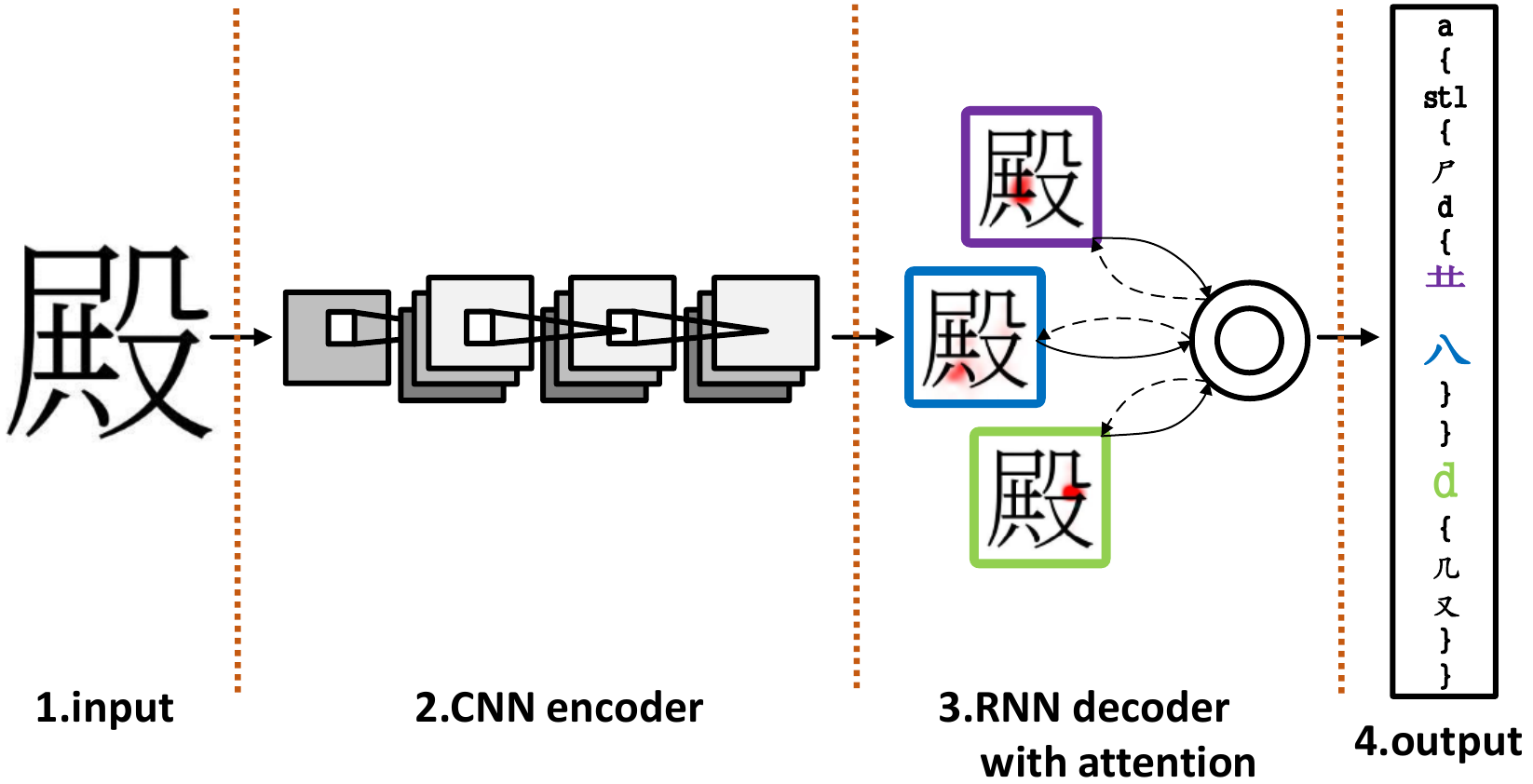}
\captionsetup{font=small}
\caption{The overall architecture of RAN for printed Chinese character recognition. The purple, blue and green rectangles are examples of attention visualization when predicting the words that are of the same color in output caption.}
\label{fig:RANarchitecture}
\end{figure}
In this paper, we evaluate RAN on printed Chinese characters. The inputs are greyscale images and the pixel value is normalized between 0 and 1. The overall architecture of RAN is shown in Fig.~\ref{fig:RANarchitecture}. We employ CNN as the encoder which is proven to be a powerful model to extract high-quality visual features from images. Rather than extracting features after a fully connected layer, we employ a CNN framework containing only convolution, pooling and activation layers, called fully convolutional neural networks. It does make sense because the subsequent decoder can selectively pay attention to certain pixels of an image by choosing specific portions from all the extracted visual features.

Assuming that the CNN encoder extracts high-level visual representations denoted by a three-dimensional array of size $H \times W \times D$, then the CNN output is a variable-length grid of \emph{L} elements, $L = H \times W$. Each of these elements is a $D$-dimensional annotation vector corresponding to a local region of the image.
\begin{equation}\label{eq:annotation}
 \mathbf{A} = \left\{ {{{\mathbf{a}}_1}, \ldots ,{{\mathbf{a}}_L}} \right\}\;,\;{{\mathbf{a}}_i} \in {\mathbb{R}^D}
\end{equation}

\subsection{Decoder with spatial attention}
\label{sec:Decoder with spatial attention}

\subsubsection{Deocder}
\label{sec:Decoder}
As illustrated in Fig.~\ref{fig:RANarchitecture}, the decoder aims to generate a caption of input Chinese character. The output caption ${\mathbf{Y}}$ is represented by a sequence of one-hot encoded words.
\begin{equation}\label{eq:outputY}
 \mathbf{Y} = \left\{ { \mathbf{y}_1, \ldots ,\mathbf{y}_C} \right\}\;,\;{{\mathbf{y}}_i} \in {\mathbb{R}^K}
\end{equation}
where $K$ is the number of total words in the vocabulary which includes the basic radicals, spatial structures and a pair of braces, $C$ is the length of caption.

Note that, the length of annotation sequence (${L}$) is fixed while the length of caption ($C$) is variable. To address the problem of associating fixed-length annotation sequences with variable-length output sequences, we attempt to compute an intermediate fixed-size vector ${{\mathbf{c}}_t}$ which will be described in Section~\ref{sec:Coverage based spatial attention}. Given the context vector ${{\mathbf{c}}_t}$, we utilize unidirectional GRU to produce captions word by word. The GRU is an improved version of simple RNN as it alleviates the problems of the vanishing gradient and the exploding gradient as described in~\cite{bengio1994learning,zhang2016rnn}. The probability of each predicted word is computed by the context vector ${{\mathbf{c}}_t}$, current GRU hidden state ${{\mathbf{s}}_t}$ and previous target word ${{\mathbf{y}}_{t - 1}}$ using the following equation:
\begin{equation}\label{eq:computePy}
  \mathbf{P}({{\mathbf{y}}_t}|{{{\mathbf{y}}_{t - 1}},\mathbf{X}}) = g \left ({{\mathbf{W}}_o} h \left({\mathbf{E}}{{\mathbf{y}}_{t - 1}} + {{\mathbf{W}}_s}{{\mathbf{s}}_t} + {{\mathbf{W}}_c}{{\mathbf{c}}_t}\right)\right )
\end{equation}
where $g$ denotes a softmax activation function over all the words in the vocabulary, $h$ denotes a maxout activation function, ${{\mathbf{W}}_o} \in {\mathbb{R}^{K \times \frac{m}{2}}}$, ${{\mathbf{W}}_s} \in {\mathbb{R}^{m \times n}}$, ${{\mathbf{W}}_c} \in {\mathbb{R}^{m \times D}}$, and ${\mathbf{E}}$ denotes the embedding matrix, $m$ and $n$ are the dimensions of embedding and GRU decoder.

The decoder adopts two unidirectional GRU layers to calculate the hidden state ${{\mathbf{s}}_t}$:
\begin{align}\label{eq:parser}
 & {{\mathbf{\hat s}}_t} = \textrm{GRU} \left( {{\bf{y}}_{t-1}}, {{\bf{s}}_{t - 1}} \right) \\
 & {\mathbf{c}_t} = f_{\text{catt}} \left( {{\mathbf{\hat s}}_t}, \bf{A} \right) \\
 & {{\mathbf{s}}_t} = \textrm{GRU} \left( {{\mathbf{c}}_t}, {{\mathbf{\hat s}}_t} \right)
\end{align}
where ${{\mathbf{s}}_{t-1}}$ denotes the previous hidden state, ${{\mathbf{\hat s}}_t}$ is the prediction of current GRU hidden state, and $f_{\text{catt}}$ denotes the coverage based spatial attention model (Section \ref{sec:Coverage based spatial attention}).

\subsubsection{Coverage based spatial attention}
\label{sec:Coverage based spatial attention}
Intuitively, for each radical or structure, the entire input character is not necessary to provide the useful information, only a part of input character should mainly contribute to the computation of context vector ${{\mathbf{c}}_t}$ at each time step $t$. Therefore, the decoder adopts a spatial attention mechanism to know which part of input is suitable to attend to generate the next predicted radical or structure and then assigns a higher weight to the corresponding local annotation vectors ${{\mathbf{a}}_i}$ (e.g. in Fig.~\ref{fig:RANarchitecture}, the higher attention probabilities are described by the lighter red color). However, there is one problem for the classic spatial attention mechanism, namely lack of coverage. Coverage means the overall alignment information that indicating whether a local part of input has been attended or not. The overall alignment information is especially important when recognizing Chinese characters because in principle, each radical or structure should be decoded only once. Lacking coverage will lead to misalignment resulting in over-parsing or under-parsing problem. Over-parsing implies that some radicals and structures have been decoded twice or more, while under-parsing denotes that some radicals and structures have never been decoded. To address this problem, we append a coverage vector to the computation of attention. The coverage vector aims at tracking the past alignment information. Here, we parameterize the coverage based attention model as a multi-layer perceptron (MLP) which is jointly trained with the encoder and the decoder:
\begin{align}\label{eq:coverage}
  & {\mathbf{F}} = {\mathbf{Q}} * \sum\nolimits_{l=1}^{t - 1} {{{\bm{\alpha}}_l}} \\
  & {e_{ti}} = {\bm{\nu }}_{\text{att}}^{\rm T}\tanh ({{\mathbf{W}}_{\text{att}}}{{\mathbf{\hat s}}_t} + {{\mathbf{U}}_{\text{att}}}{{\mathbf{a}}_i} + {{\mathbf{U}}_f}{{\mathbf{f}}_i}) \\
  & {\alpha _{ti}} = \frac{{\exp ({e_{ti}})}}{{\sum\nolimits_{k = 1}^L {\exp ({e_{tk}})} }}
\end{align}
where ${e_{ti}}$ denotes the energy of annotation vector ${{\mathbf{a}}_{i}}$ at time step $t$ conditioned on the current GRU hidden state prediction ${{\mathbf{\hat s}}_t}$ and coverage vector ${{\mathbf{f}}_i}$. The coverage vector is initialized as a zero vector and we compute it based on the summation of all past attention probabilities. ${\alpha _{ti}}$ denotes the spatial attention coefficient of ${{\mathbf{a}}_{i}}$ at time step $t$. Let $n'$ denote the attention dimension and $M$ denote the number of feature maps of filter $\mathbf{Q}$; then ${{\bm{\nu }}_{\text{att}}} \in {\mathbb{R}^{{n'}}}$, ${{\mathbf{W}}_{\text{att}}} \in {\mathbb{R}^{{n'} \times n}}$, ${{\mathbf{U}}_{\text{att}}} \in {\mathbb{R}^{{n'} \times D}}$ and ${{\mathbf{U}}_{f}} \in {\mathbb{R}^{{n'} \times M}}$.
With the weights ${\alpha _{ti}}$, we compute the context vector ${{\mathbf{c}}_t}$ as:
\begin{equation}\label{eq:context vector}
  {{\mathbf{c}}_t} = \sum\nolimits_{i=1}^L {{\alpha _{ti}}{{\mathbf{a}}_i}}
\end{equation}
which has a fixed-length one regardless of input image size.

\begin{figure}
\centering
\includegraphics[width=3in]{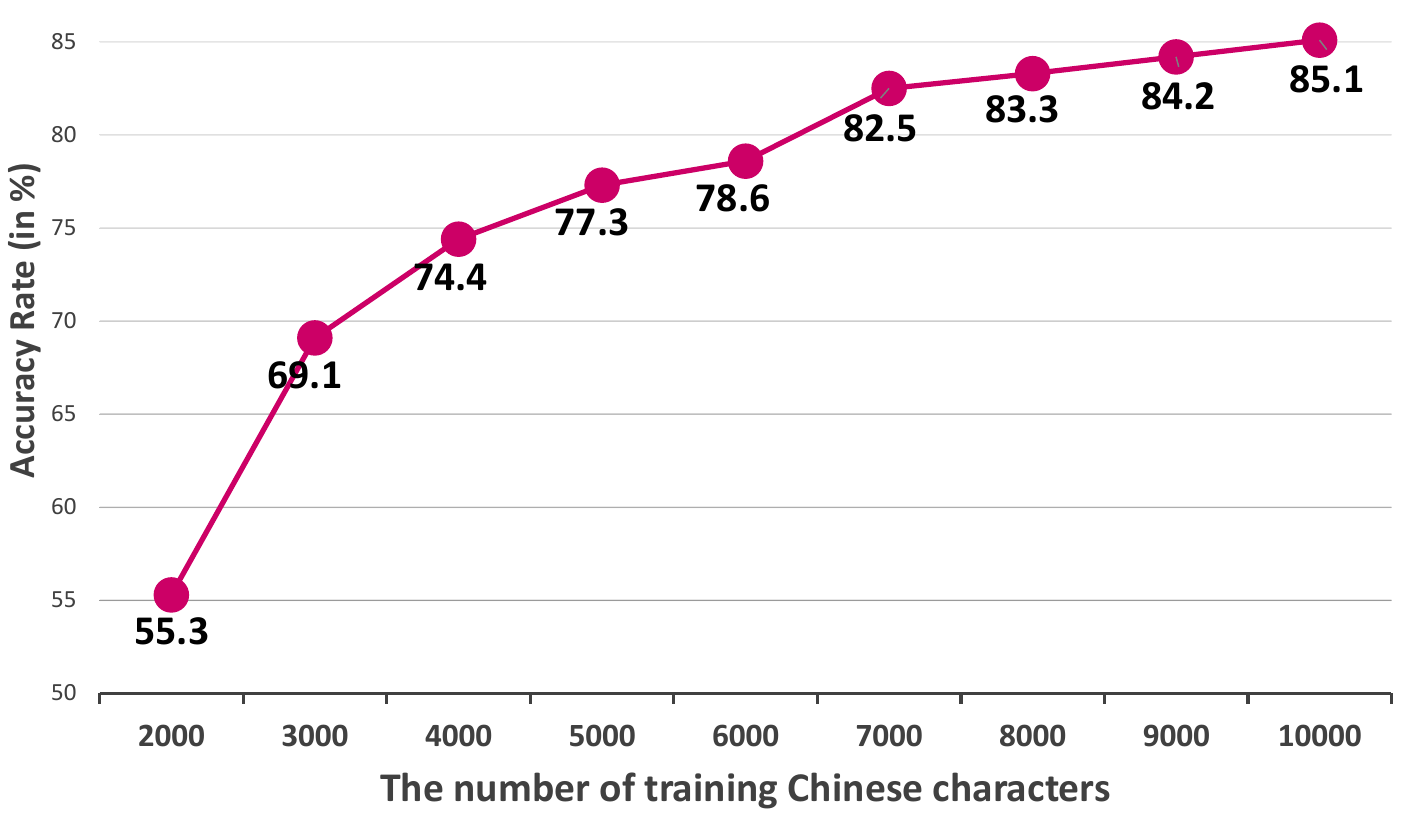}
\captionsetup{font=small}
\caption{The performance comparison of RAN for recognizing unseen Chinese characters with respect to the number of training Chinese characters.}
\label{fig:0shotExperiments}
\end{figure}
\section{Experiments}
\label{sec:Experiments}
The training objective of RAN is to maximize the predicted word probability in Eq.~\eqref{eq:computePy} and we use cross-entropy (CE) as the loss function. The CNN encoder employs VGG~\cite{simonyan2014very} architecture. We propose to use two different VGG architectures, namely VGG14-s and VGG14, due to the different size of training set. The "VGG14" means that the two architectures are both composed of 14 convolutional layers, which are divided into 4 blocks. The number of stacked convolutional layers in each block is (3, 3, 4, 4) and a max-pooling function is followed after each block. In VGG14-s, the number of output channels in each block is (32, 64, 128, 256). While in VGG14, the number of output channels in each block is (64, 128, 256, 512). Note that, the number of output channels of each convolutional layer in the same block remains unchanged. We use VGG14-s for experiments of recognizing unseen characters because the size of training set is relatively small, the suffix "s" means "smaller". While the VGG14 is used for experiments of recognizing seen characters with various font styles.
The decoder is a single layer with 256 forward GRU units. The embedding dimension $m$ and GRU decoder dimension $n$ are set to 256. The attention dimension $n'$ is set to the annotation dimension $D$. The convolution kernel size for computing coverage vector is set to (5 $\times$ 5) and the number of feature maps $M$ is set to 256. We utilize the adadelta algorithm~\cite{zeiler2012adadelta} with gradient clipping for optimization.

In the decoding stage, we aim to generate a most likely caption string given the input character. The beam search algorithm~\cite{zhang2017watch} is employed to complete the decoding process because we do not have the ground-truth of previous predicted word during testing. The beam size is set to 10.

\subsection{Experiments on recognition of unseen characters}
\label{sec:performance on recognition of unseen characters}

\begin{figure}
\centering
\includegraphics[width=2.5in]{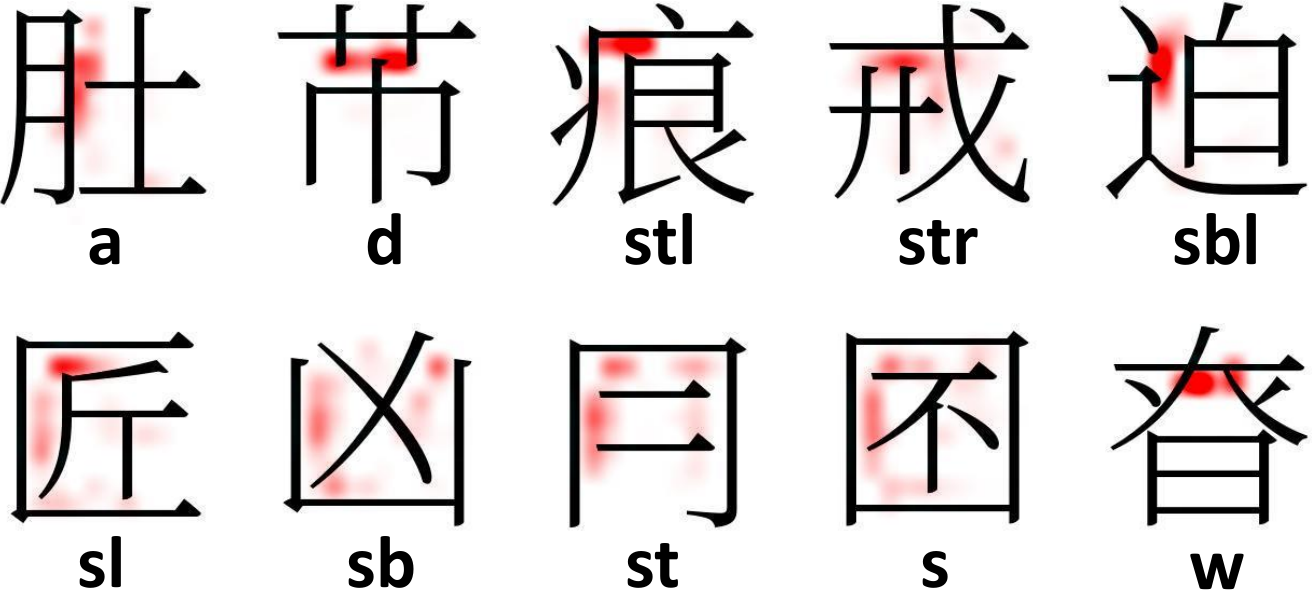}
\captionsetup{font=small}
\caption{Attention visualization of identifying ten common structures of Chinese characters.}
\label{fig:StructureAttention}
\end{figure}

\begin{figure}
\centering
\includegraphics[width=3.4in]{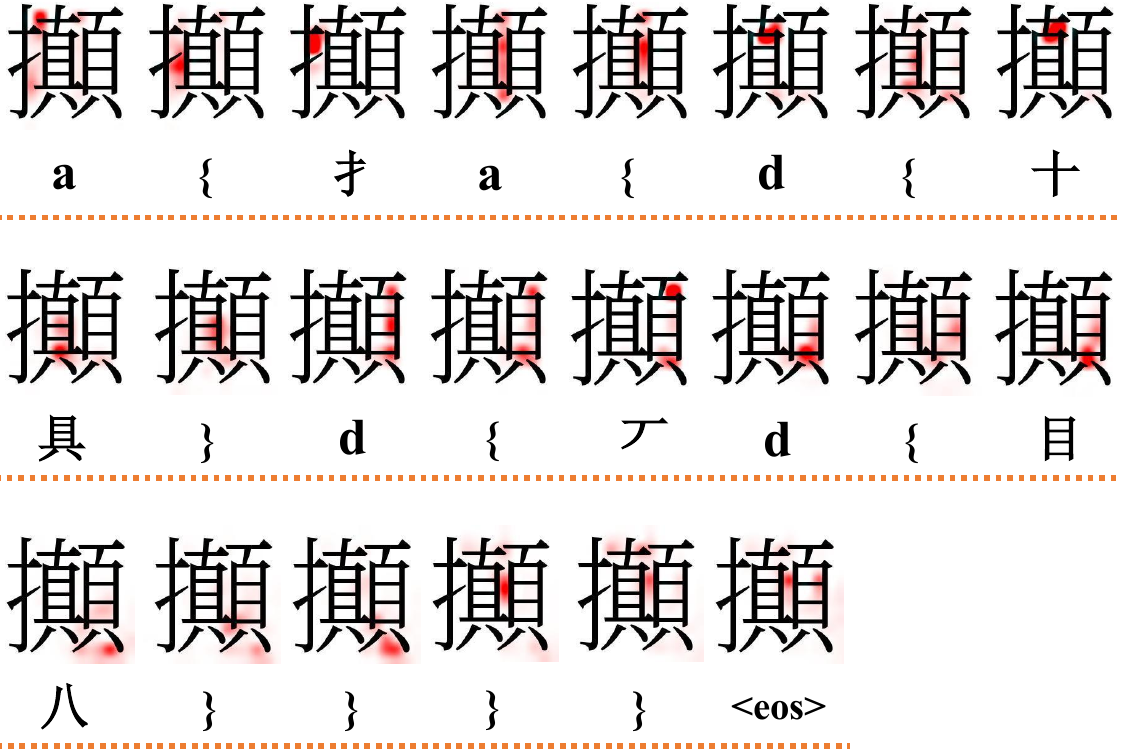}
\captionsetup{font=small}
\caption{Attention visualization of recognizing an unseen Chinese character.}
\label{fig:AttentionExample}
\end{figure}

In this section, we show the effectiveness of RAN on identifying unseen Chinese characters through accuracy rate and attention visualization. A test character is considered as successfully recognized only when its predicted caption exactly matches the ground-truth.

In this experiment, we choose 26,079 Chinese characters in Song font style which are composed of only 361 radicals and 29 spatial structures. We divide them into training set, validation set and testing set. We increase the training set from 2,000 to 10,000 Chinese characters to see how many training characters are enough to train our model to recognize the unseen 16,079 characters. As for the unseen 16,079 Chinese characters, we choose 2,000 characters as the validation set and 14,079 characters as the testing set. We also employ an ensemble method during testing procedure~\cite{zhang2017watch} because the performances vary severely due to the small training set. We illustrate the performance in Fig.~\ref{fig:0shotExperiments}, where 2,000 training Chinese characters can successfully recognize 55.3\% unseen 14,079 Chinese characters and 10,000 training Chinese characters can successfully recognize 85.1\% unseen 14,079 Chinese characters. Actually, only about 500 Chinese characters are adequate to cover overall Chinese radicals and spatial structures. However, our experiments start from 2,000 training characters because RAN is hard to converge when the training set is too small.

To generate a Chinese character caption, it is essential to identify the structures between isolated radicals. As illustrated in Fig.~\ref{fig:StructureAttention}, we show ten examples on how RAN identifies common structures through attention visualization. The red color in attention maps represents the spatial attention probabilities, where the lighter color describes the higher attention probabilities and the darker color describes the lower attention probabilities. Taking ``a'' structure as an example, the decoder mainly focuses on the blank region between two radicals, indicating a left-right direction.

More specifically, in Fig.~\ref{fig:AttentionExample}, we show that how RAN learns to recognize an unseen Chinese character from an image into a character caption step by step. When encountering basic radicals, the attention model well generates the alignment strongly corresponding to the human intuition. Also, it successfully generates the structure ``a'' and ``d'' when it detects a left-right direction and a top-bottom direction. Immediately after detecting a spatial structure, the decoder generates a pair of braces ``\{\}'', which is employed to constrain the structure in Chinese character caption.

\subsection{Experiments on recognition of seen characters}
\label{sec:performance on recognition of seen characters}
\begin{figure}
\centering
\includegraphics[width=3.4in]{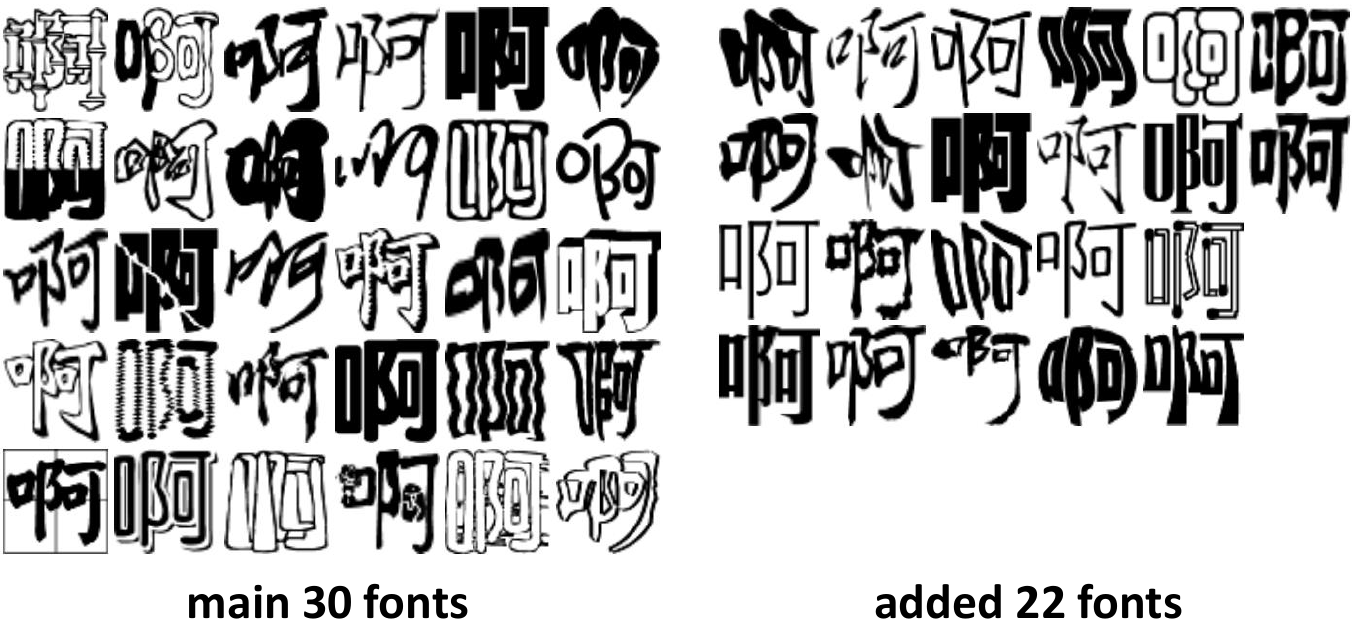}
\captionsetup{font=small}
\caption{Visualization of font styles.}
\label{fig:fontstyles}
\end{figure}

In this section, we show the effectiveness of RAN on recognition of seen Chinese characters by comparing it with traditional whole-character modeling methods.
\begin{figure}
\centering
\includegraphics[width=3.4in]{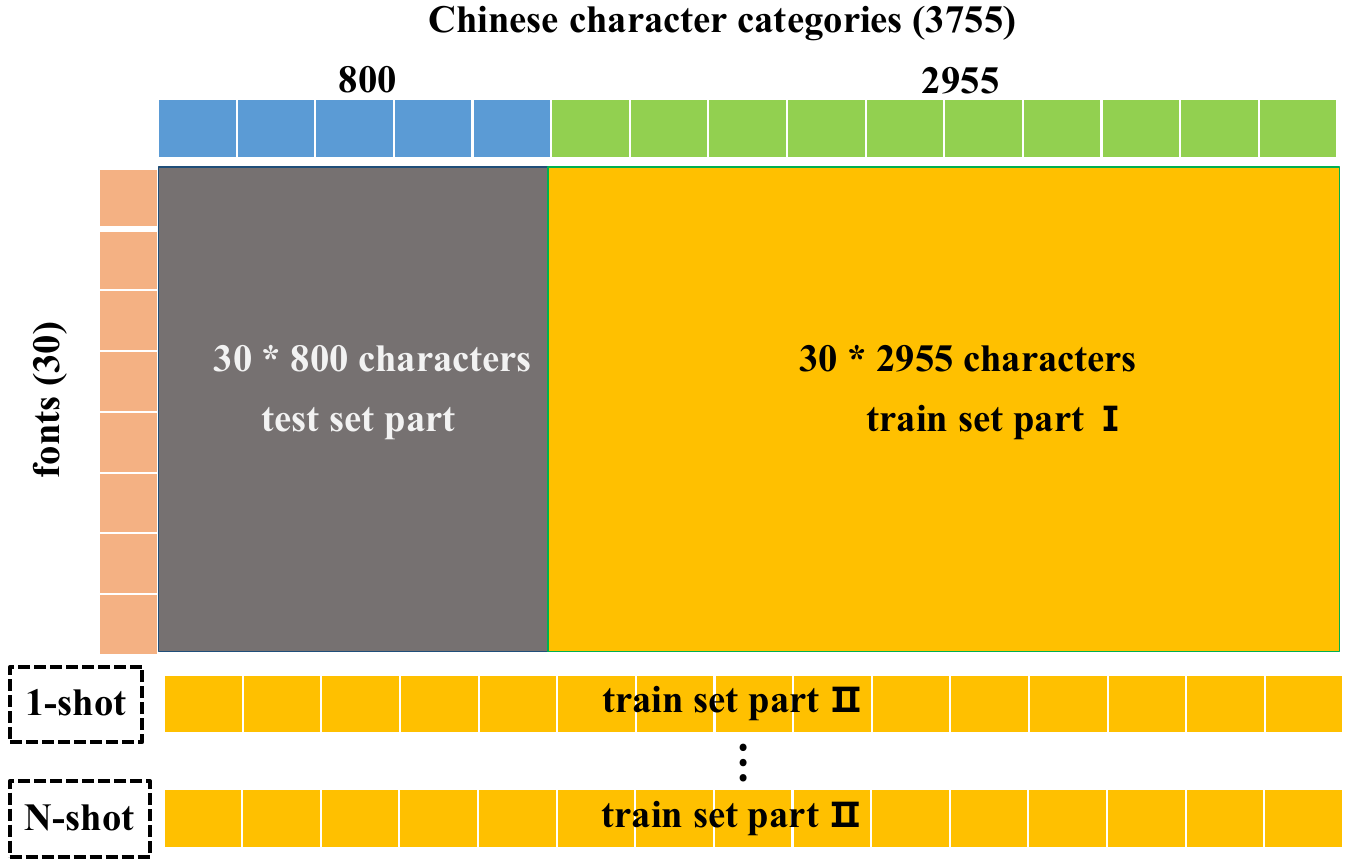}
\captionsetup{font=small}
\caption{Description of dividing training set and testing set for experiments of recognizing seen Chinese characters.}
\label{fig:OneshotDataset}
\end{figure}
The training set contains 3755 common used Chinese character categories and the testing set contains 800 character categories. The detailed implementation of organizing training set and testing set is illustrated in Fig.~\ref{fig:OneshotDataset}. We design this experiment like few-shot learning of Chinese character recognition. We divide 3755 characters into 2955 characters and other 800 characters. The 800 characters with 30 various font styles form the testing set and the other 2955 characters with the same 30 font styles become a part of training set. Additionally, we use 3755 characters with other font styles as a second part of training set. When we add 3755 characters with other N font styles as the second training set, we call this experiment N-shot. The number of font styles of the second part of training set increased from 1 to 22. The description of main 30 font styles and newly added 22 font styles are visualized in Fig.~\ref{fig:fontstyles}.

\begin{table}[t]
\captionsetup{font=small}
\caption{\label{tab:oneshot results}{Comparison of Accuracy Rate (AR in \%) between RAN with traditional whole-character based approaches.}}
\centering
\begin{tabular}{c c c c c}
\toprule
\textbf{} & \textbf{1-shot} & \textbf{2-shot} & \textbf{3-shot} & \textbf{4-shot}\\
\midrule
Zhong \cite{zhong2015multi} & 4 & 16.3 & 43.7 & 62.4 \\
VGG14 & 23.4 & 58.4 & 74.3 & 82.2 \\
\textbf{RAN} & \textbf{80.7} & \textbf{85.7} & \textbf{87.9} & \textbf{88.3} \\
\bottomrule
\end{tabular}
\end{table}

\begin{figure}
\centering
\includegraphics[width=3.5in]{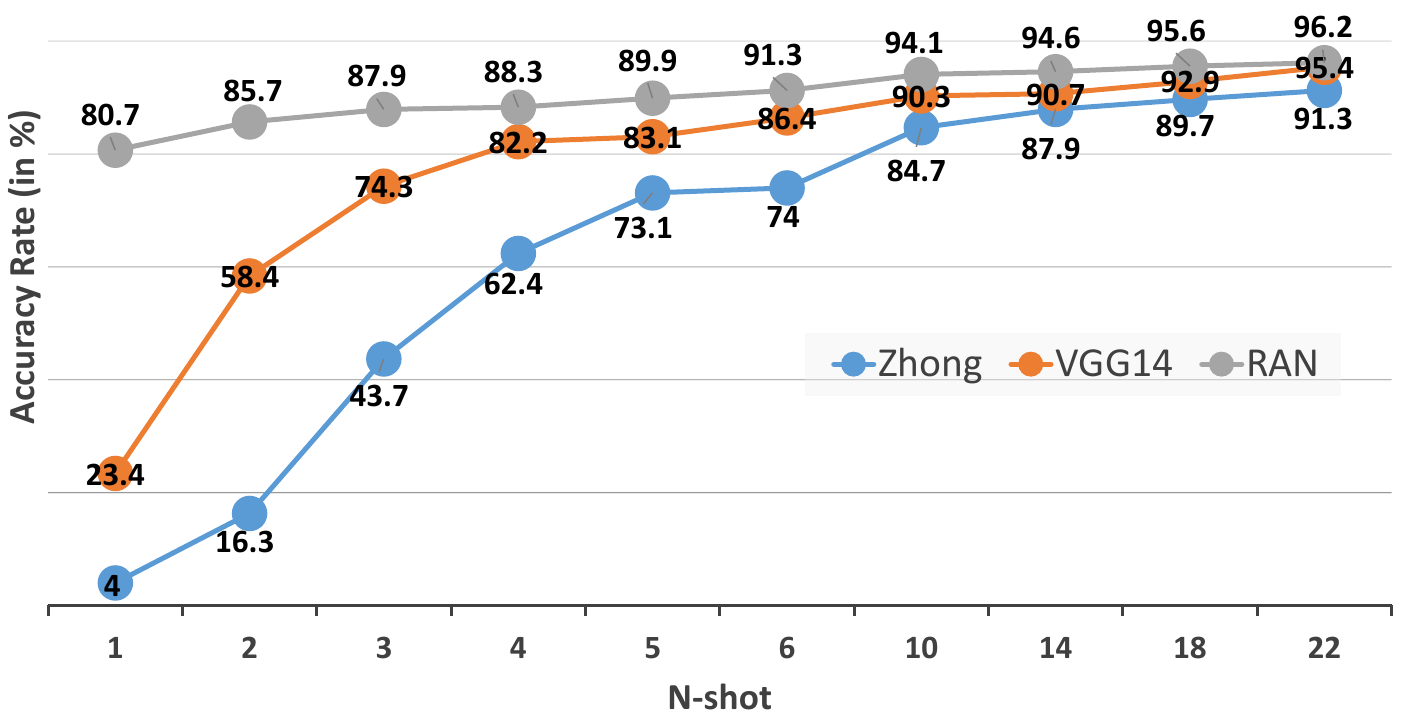}
\captionsetup{font=small}
\caption{The performance comparison among Zhong, VGG14 and RAN with respect to the number of newly added training fonts N.}
\label{fig:NshotExperiments}
\end{figure}

The system ``Zhong'' is the proposed system in~\cite{zhong2015multi}. Also, we replace the CNN architecture in ``Zhong'' with VGG14 and keep the other parts unchanged, we name it as system ``VGG14''. Table~\ref{tab:oneshot results} clearly declares that RAN significantly outperforms the traditional whole-character based approaches on seen Chinese characters, when the training samples of these character classes are very few. Fig.~\ref{fig:NshotExperiments} illustrates the comparison when the number of training samples of seen classes increases and RAN still consistently outperforms whole-character based approaches.

\section{Conclusions and future work}
\label{sec:Conclusion and future work}

In this paper, we introduce a novel model named radical analysis network for zero-shot learning of Chinese characters recognition. We show from experimental results that RAN is capable of recognizing unseen Chinese characters with visualization of spatial attention and outperforms traditional whole-character based approaches on seen Chinese character recognition.
In future work, we plan to investigate RAN's ability in recognizing handwriting Chinese characters or Chinese characters in natural scene. We will also explore the effect of mapping relationship between structure/radical and Chinese character in improving the performance of RAN.

\section{Acknowledgments}
This work was supported in part by the National Key R\&D Program of China under contract No. 2017YFB1002202, the National Natural Science Foundation of China under Grants No. 61671422 and U1613211, the Key Science and Technology Project of Anhui Province under Grant No. 17030901005, and MOE-Microsoft Key Laboratory of USTC.


\ninept
\bibliographystyle{IEEEbib}
\bibliography{RAN-icme2018}

\end{document}